\documentclass[10pt,twocolumn,letterpaper]{article}

\usepackage{times}
\usepackage{epsfig}
\usepackage{graphicx}
\usepackage{amsmath}
\usepackage{amssymb}
\usepackage{makecell}
\usepackage{xcolor}
\usepackage[utf8]{inputenc} 
\usepackage[T1]{fontenc}    
\usepackage{url}            
\usepackage{booktabs}       
\usepackage{amsfonts}       
\usepackage{nicefrac}       
\usepackage{microtype}      
\usepackage{lipsum}
\usepackage[bookmarks=false]{hyperref}

\usepackage{arydshln}
\usepackage{multirow}
\usepackage{floatpag}
\usepackage{multicol}
\usepackage{tikz}

\def\etc{\textit{etc.~}}
\def\etal{\textit{et al. }}
\def\ourmodel{BSUV-Net}
\newcommand{\basealgo}[1]{self-contained algorithms}
\newcommand{\Basealgo}[1]{Self-contained algorithms}

\usepackage{pifont}
\newcommand{\cmark}{\ding{51}}%
\def\new#1{{#1}}

\begin{document}

\title{ BSUV-Net: A Fully-Convolutional Neural Network for\\ Background Subtraction of Unseen Videos}

\author{
  M. Ozan Tezcan, Prakash Ishwar, Janusz Konrad
    \thanks{This work was supported in part by ARPA-E under agreement DE-AR0000944 and by the donation of Titan GPUs from NVIDIA Corp.}\\\\
  Department of Electrical and Computer Engineering\\
  Boston University, Boston, MA \\
  \texttt{[mtezcan, pi, jkonrad]@bu.edu} \\
}

\date{}
\maketitle


\begin{abstract}
   Background subtraction is a basic task in computer vision and video processing often applied as a pre-processing step for object tracking, people recognition, \etc Recently, a number of successful background-subtraction algorithms have been proposed, however nearly all of the top-performing ones are supervised. {Crucially, their success relies upon} the availability of some {\it annotated} frames of the test video {during training.} Consequently, their performance on completely ``unseen'' videos {is undocumented in the literature.} In this work, we {propose} a new, supervised, background-subtraction algorithm for {\textit{unseen videos}} ({\ourmodel}) based on a fully-convolutional neural network. The input to our network consists of the current frame and two background frames captured at different time scales along with their semantic segmentation maps. In order to reduce the chance of overfitting, we {also introduce} a new data-augmentation technique which {mitigates the impact of} illumination difference between the background frames and {the current} frame. \new{On the CDNet-2014 dataset, {\ourmodel} outperforms state-of-the-art algorithms evaluated on unseen videos in terms of several metrics including F-measure, recall and precision.}
\end{abstract}
\vspace{-2ex}
\section{Introduction}
\label{sec:intro}
Background subtraction (BGS) is
a foundational, low-level task in computer vision and video processing. The aim of BGS is to segment an input video frame into regions corresponding to either foreground (e.g., motor vehicles) or background (e.g., highway surface).
It is frequently used as a pre-processing step for higher-level tasks such as object tracking, people and motor-vehicle recognition, human activity recognition, etc. Since BGS is often the first pre-processing step, the accuracy of its output has an overwhelming impact on the overall performance of subsequent steps. Therefore, it is critical that BGS produce as accurate a foreground/background segmentation as possible.

Traditional BGS algorithms are unsupervised and rely on a background model to predict foreground regions \cite{stauffer1999gmm, zivkovic2004improvedgmm, elgammal2002kde, mittal2004adaptivekde, barnich2011vibe, st2015pawcs, st2015subsense, icsik2018swcd, lee2018wisenetmd}. PAWCS \cite{st2015pawcs}, SWCD \cite{icsik2018swcd} and WisenetMD \cite{lee2018wisenetmd} are considered to be state-of-the-art unsupervised BGS algorithms. However, since they rely on the accuracy of the background model, they encounter difficulties when applied to complex scenes. Recently, ensemble methods and a method leveraging semantic segmentation have been proposed which significantly outperform traditional algorithms \cite{bianco2017iutis, zeng2019cnnsfc, braham2017semanticbgs}.

The success of deep learning in computer vision did not bypass BGS research \cite{bouwmans2019deep}. A number of {\it supervised} deep-learning BGS algorithms have been developed \cite{braham2016deepbgs, babaee2018cnnbgs, bakkay2018bscgan, zeng2018mfcnnbgs, wang2017interactivebgs, lim2018fgsegnet, lim2018fgsegnetv3,sakkos20173dbgs} with performance easily surpassing that of traditional methods. However, most of these algorithms have been tuned to either one specific video or to a group of similar videos, and their performance on unseen videos has not been evaluated. For example, FgSegNet \cite{lim2018fgsegnet} uses 200 frames from a test video for training and the remaining frames from the {\it same} video for evaluation. If applied to an unseen video, its performance drops significantly (Section~\ref{sec:results}).

In this paper, we introduce {\it Background Subtraction for Unseen Videos} (\ourmodel), a fully-convolutional neural network for predicting foreground of an {\it unseen} video. A key feature of our approach is that the training and test sets are composed of frames originating from different videos.
This guarantees that no ground-truth data from the test videos have been shown to the network in the training phase.
By employing two reference backgrounds at different time scales, {\ourmodel} addresses two challenges often encountered in BGS: varying scene illumination and intermittently-static objects that tend to get absorbed into the background.
We also propose novel data augmentation which further improves our method's performance under varying illumination.
Furthermore, motivated by recent work on the use of semantic segmentation in BGS \cite{braham2017semanticbgs}, we improve our method's accuracy by inputting semantic information along with the reference backgrounds and current frame.
The main contributions of our work are as follows:
\begin{enumerate} \topsep -2pt\partopsep -2pt\itemsep -2pt
    \item {\textbf{Supervised BGS for Unseen Videos:}} Although supervised algorithms, especially neural networks, have significantly improved BGS performance, they are tuned to a specific video and thus their performance on {\it unseen} videos deteriorates dramatically. To the best of our knowledge, {\ourmodel} is the first supervised BGS algorithm that is truly generalizable to {\it unseen} videos.
    \item {\textbf{Data Augmentation for Increased Resilience to Varying Illumination:}} 
    Changes in scene illumination pose a major challenge to BGS algorithms. 
    To mitigate this, we develop a simple, yet effective, data augmentation technique.
    Using a simple additive model we vary the illumination of
    the current frame and the reference background frames that are fed into {\ourmodel} during training. This enables us to effectively tackle various illumination change scenarios that may be present in test videos.
    \item \textbf{Leveraging Semantic and Multiple Time-Scale Information:} {\ourmodel} improves foreground-boundary segmentation accuracy by accepting semantic information as one of its inputs. This is unlike in an earlier BGS method \cite{braham2017semanticbgs} which used semantic information as a \textit{post-processing} step. The other network inputs are the current frame (to be segmented) and a two-frame background model with data from different time scales. While one background frame, based on {\it distant} history, helps with the discovery of intermittently-static objects, the other frame, based on {\it recent} history, is key for handling dynamic factors such as illumination changes.
\end{enumerate}

Based on our extensive experiments on the CDNet-2014 dataset \cite{goyette2012changedetection}, {\ourmodel} outperforms state-of-the-art BGS algorithms evaluated on {\it unseen} videos

\vspace{-1ex}
\section{Related Work}
\label{sec:rel_work}

A wide range of BGS algorithms have been developed in the past,
each having some advantages and disadvantages over others. 
Since this is not a survey paper, we will not cover all BGS variants. Instead, we will focus only on 
recent top-performing methods.
We divide these algorithms into 3 categories: (i) BGS by (unsupervised) background modeling, (ii) supervised BGS tuned to a single video or a group of videos, (iii) Improving BGS algorithms by post-processing.

\subsection{BGS by Background Modeling}

Nearly all traditional BGS algorithms first compute a background model, and then use it to predict the foreground. While a simple model based on the mean or median of a subset of preceding frames offers only a single background value per pixel, a probabilistic Gaussian Mixture Model (GMM) \cite{stauffer1999gmm} allows a range of background values. This idea was improved by creating an online procedure for the update of GMM parameters in a pixel-wise manner \cite{zivkovic2004improvedgmm}. Kernel Density Estimation (KDE) was introduced into BGS \cite{elgammal2002kde} as a non-parametric alternative to GMMs and was subsequently improved \cite{mittal2004adaptivekde}. The probabilistic methods achieve better performance compared to single-value models for dynamic scenes and scenes with small background changes.

In \cite{barnich2011vibe}, Barnich and Droogenbroeck introduced a sample-based background model. Instead of implementing a probabilistic model, they modeled the background by a set of sample values per pixel and used a distance-based model to decide whether a pixel should be classified as background or foreground. Since color information alone is not sufficient for complex cases, such as illumination changes, Bilodeau \etal introduced Local Binary Similarity Patterns (LBSP) to compare the current frame and background using spatio-temporal features instead of color \cite{bilodeau2013lbsp}. St-Charles \etal combined color and texture information, and introduced a word-based approach, PAWCS \cite{st2015pawcs}. They considered pixels as background words and updated each word's reliability by its persistence. Similarly, SuBSENSE by St-Charles \etal \cite{st2015subsense} combines LBSP and color features, and employs pixel-level feedback to improve the background model. 
    
Recently, Isik \etal introduced SWCD, a pixel-wise, sliding-window approach leveraging a dynamic control system to update the background model \cite{icsik2018swcd}, while Lee \etal introduced WisenetMD, a multi-step algorithm to eliminate false positives in dynamic backgrounds \cite{lee2018wisenetmd}. 
In \cite{sultana2019unsupervised}, Sultana \etal introduced an unsupervised background estimation method based on a generative adversarial network (GAN).
They use optical flow to create a motion mask and then in-paint covered regions with background values estimated by a GAN. The foreground is then computed by subtracting the estimated background from the current frame followed by morphological operations. 
They, however, do not achieve state-of-the-art results.
Zeng \etal introduced RTSS \cite{zeng2019rtss} which uses deep learning-based semantic segmentation predictions to improve the background model used in SubSENSE \cite{st2015subsense}.

\subsection{Supervised BGS}
\label{sec:supervisedbgs}

Although background subtraction has been extensively studied in the past, the definition of a supervised BGS algorithm is still vague. Generally speaking, the aim of a supervised BGS algorithm is to learn the parameters (e.g., neural-network weights) of a complex function in order to minimize a loss function of the labeled training frames. Then, the performance of the algorithm is evaluated on a separate set of test frames. In this section we divide the supervised BGS algorithms into three groups namely, \textit{video-optimized}, \textit{video-group-optimized} and \textit{video-agnostic} depending on which frames and videos they use during training and testing. 

Several algorithms use some frames from a test video for training and all the frames of the {\it same} video for evaluating performance on that video. In such algorithms, parameter values are optimized separately for each video. We will refer to this class of algorithms as \textit{video-optimized} BGS algorithms.  In another family of algorithms, randomly-selected frames from a \textit{group} of test videos are used for training and all the frames of the same videos are used for testing. Since some frames from {\it all} test videos are used for training, we will refer this class of algorithms as \textit{video-group-optimized} algorithms. Note that, in both of these scenarios the algorithms are neither optimized for nor evaluated on {\it unseen} videos and to the best of our knowledge all of the top-performing supervised BGS algorithms to-date are either \textit{video-optimized} or \textit{video-group-optimized}. In this paper, we introduce a new category of supervised BGS algorithms, called \textit{video-agnostic} algorithms, that can be applied to unseen videos with no or little loss of performance. To learn parameters, a {\it video-agnostic} algorithm uses frames from a set of training videos but for performance evaluation it uses a completely different set of videos.

In recent years, supervised learning algorithms based on convolutional neural networks (CNNs) have been widely applied to BGS. The first CNN-based BGS algorithm was introduced in \cite{braham2016deepbgs}. This is a \textit{video-optimized} algorithm which produces a single foreground probability for the center of each $27 \times 27$ patch of pixels.
A method proposed in \cite{wang2017interactivebgs} uses a similar approach, but with a modified CNN which operates on patches of size $31 \times 31$ pixels.
    
Instead of using a patch-wise algorithm, Zeng and Zhu introduced the Multiscale Fully-Convolutional Neural Network (MFCN) which can predict the foreground of the entire input image frame in one step \cite{zeng2018mfcnnbgs}. Lim and Keles proposed a triplet CNN which uses siamese networks to create features at three different scales and combines these features within a transposed CNN \cite{lim2018fgsegnet}. In a follow-up work, they removed the triplet networks and used dilated convolutions to capture the multiscale information \cite{lim2018fgsegnetv3}. In \cite{bakkay2018bscgan}, Bakkay \etal used generative adversarial networks for BGS. The generator performs the BGS task, whereas the discriminator tries to classify the BGS map as real or fake. Although all these algorithms perform very well on various BGS datasets, it is important to note that they are all \textit{video-optimized}, thus they will suffer a performance loss when tested on unseen videos. In \cite{babaee2018cnnbgs}, Babae \etal designed a \textit{video-group-optimized} CNN for BGS. They randomly selected $5\%$ of CDNet-2014 frames \cite{goyette2012changedetection}  as a training set and developed a single network for all of the videos in this dataset. In \cite{sakkos20173dbgs}, Sakkos \etal used a 3D CNN to capture the temporal information in addition to the color information. Similarly to \cite{babaee2018cnnbgs}, they trained a single algorithm using 70\% of frames in CDNet-2014 and then used it to predict the foreground in all videos of the dataset. Note that even these approaches do not  generalize to other videos since some ground truth data from each video exists in the training set. Table \ref{table:training_schemes} compares and summarizes the landscape of supervised BGS algorithms and the methodology used for training and evaluation.
    
As discussed above, none of the CNN-based BGS algorithms to-date have been designed for or tested on unseen videos with no ground truth at all. 
This limits their practical utility since it is not possible to label some frames in {\it every} new video. 
Since the publication of the first version of this paper, we have learned about a recent BGS algorithm named $3DFR$ \cite{mandal20193dfr}, which uses $3D$ spatio-temporal convolution blocks in an encoder-decoder architecture to predict background in an unseen video. However, \cite{mandal20193dfr} only reports evaluation results on 10 out of the 53 videos of CDNet2014.

\begin{table*}
\centering
\caption{Training/evaluation methodologies of supervised BGS algorithms on CDNet-2014.}
\begin{tabular}{| c | c | c | c|}
    \hline
     \textbf{Algorithm} & \multicolumn{2}{c|}{\textbf{Are some frames from test videos used in training?}} & \makecell{\textbf{Training and evaluation} \\ \textbf{methodology}}\\
     \hline
     Braham-CNN-BGS \cite{braham2016deepbgs} & Yes & First half of the labeled  frames of the test video & \textit{video-optimized}\\
     \hline
     MFCNN \cite{zeng2018mfcnnbgs} & Yes & \makecell{Randomly selected 200 frames from the first \\ 3000 labeled frames of the test video} & \textit{video-optimized}\\
     \hline
     \makecell{Wang-CNN-BGS \cite{wang2017interactivebgs} \\ FGSegNet \cite{lim2018fgsegnet, lim2018fgsegnetv3} \\ BScGAN \cite{bakkay2018bscgan}} & Yes & Hand picked 200 labeled frames of the test video & \textit{video-optimized}\\
     \hline
     Babae-CNN-BGS \cite{babaee2018cnnbgs} & Yes & $5\%$ of the labeled frames of all videos & \textit{video-group-optimized}\\
     \hline
     3D-CNN-BGS \cite{sakkos20173dbgs} & Yes & $70\%$ of the labeled frames of all videos & \textit{video-group-optimized}\\
     \hline 
     {\bf\ourmodel} (proposed) & No & No frame from test videos is used in training & \textit{video-agnostic}\\
     \hline 
\end{tabular}
\label{table:training_schemes}
\end{table*}

\subsection{Improving BGS Algorithms by Post-Processing}
    
Over the last few years, many deep-learning-based algorithms were developed for the problem of semantic segmentation and they achieved state-of-the-art performance. In \cite{braham2017semanticbgs}, Braham and Droogenbroeck introduced a post-processing step for BGS algorithms based on semantic segmentation predictions. Given an input frame, they predicted a segmentation map using PSPNet \cite{zhao2017pspnet} and obtained pixel-wise probability predictions for semantic labels such as person, car, animal, house etc. Then, they manually grouped these labels into two sets -- foreground and background labels,
and used this information to improve any BGS algorithm's output in a post-processing step. They obtained very competitive results 
by using SubSENSE \cite{st2015subsense} as the BGS algorithm. 

Bianco \etal introduced an algorithm called IUTIS which combines the results produced by several BGS algorithms \cite{bianco2017iutis}. They used genetic programming to determine how to combine several BGS algorithms' outputs using a sequence of basic binary operations, such as logical ``and/or'', majority voting and median filtering. Their best result was achieved by using 5 top-performing BGS algorithms on the CDNet-2014 dataset at the time of publication. Zeng \etal followed the same idea, but instead of genetic programming, used a fully-convolutional neural network to fuse several BGS results into a single output \cite{zeng2019cnnsfc}, and outperformed IUTIS on CDNet-2014.

\section{Proposed Algorithm: \ourmodel}
\label{sec:algo}
    
\subsection{Inputs to {\ourmodel}}
\label{sec:input}
    
Segmenting an unseen video frame into foreground and background regions without using any information about the background would be an ill-defined problem.
%
%
In {\ourmodel}, we use two reference frames to characterize the background. One frame is an ``empty'' background frame, with no people or other objects of interest, which can typically be extracted from the beginning of a video e.g., {\it via} median filtering over a large number of \textit{initial} frames. This provides an accurate reference that is very helpful for segmenting intermittently-static objects in the foreground. However, due to dynamic factors, such as illumination variations, this reference may not be valid after some time. To counteract this, we use another reference frame that characterizes \textit{recent} background, for example by computing median of 100 frames preceding the frame being processed. However, this frame might not be as accurate as the first reference frame since we cannot guarantee that there will be no foreground objects in it (if such objects are present for less than 50 frames, the temporal median will suppress them). By using two reference frames captured at different time scales, we aim to leverage benefits of each frame type.
   
Braham \etal \cite{braham2017semanticbgs} have shown that leveraging results of  semantic segmentation significantly improves the performance of a BGS algorithm, for example by using semantic segmentation results in a post-processing step. In {\ourmodel}, we follow a different idea and use semantic information as an additional input channel to our neural network. In this way, we let our network learn how to use this information. To extract semantic segmentation information, we used a state-of-the-art CNN called DeepLabv3 \cite{chen2017deeplab} trained on ADE20K \cite{zhou2017ade20k}, an extensive semantic-segmentation dataset with 150 different class labels and more than 20,000 images with dense annotations. Let us denote the set of object classes in ADE20K as $C = \{c_0,\ c_1,\ \dots, \ c_{149}\}$. Following the same procedure as in \cite{braham2017semanticbgs}, we divided these classes into two sets: foreground and background objects. As foreground objects, we used person, car, cushion, box, book, boat, bus, truck, bottle, van, bag and bicycle. 
The rest of the classes are used as background objects. The softmax layer of DeepLabv3 provides pixel-wise class probabilities $p_{c_j}$ for $c_j\in C$. Let ${\mathbf{I}}[m,n]$ be an input frame at spatial location $m,n$ and let $\{p_{c_j}[m, n]\}_{j=0}^{149}$ be the predicted probability distribution of ${\mathbf{I}}[m,n]$. We compute a foreground probability map (FPM) $\mathbf{S}[m, n] = \sum_{{c_j} \in F} p_{c_j}[m, n]$, where $F$ is the set of foreground classes.

We use the current, recent and empty frames in color, each along with its FPM, as the input to BSUV-Net (Figure~\ref{fig:network}). Clearly, the number of channels in {\ourmodel}'s input layer is 12 for each frame consists of 4 channels (R,G,B,FPM).

\subsection{Network Architecture and Loss Function}
    
\begin{figure*}
  \centering
  \includegraphics[width=17cm]{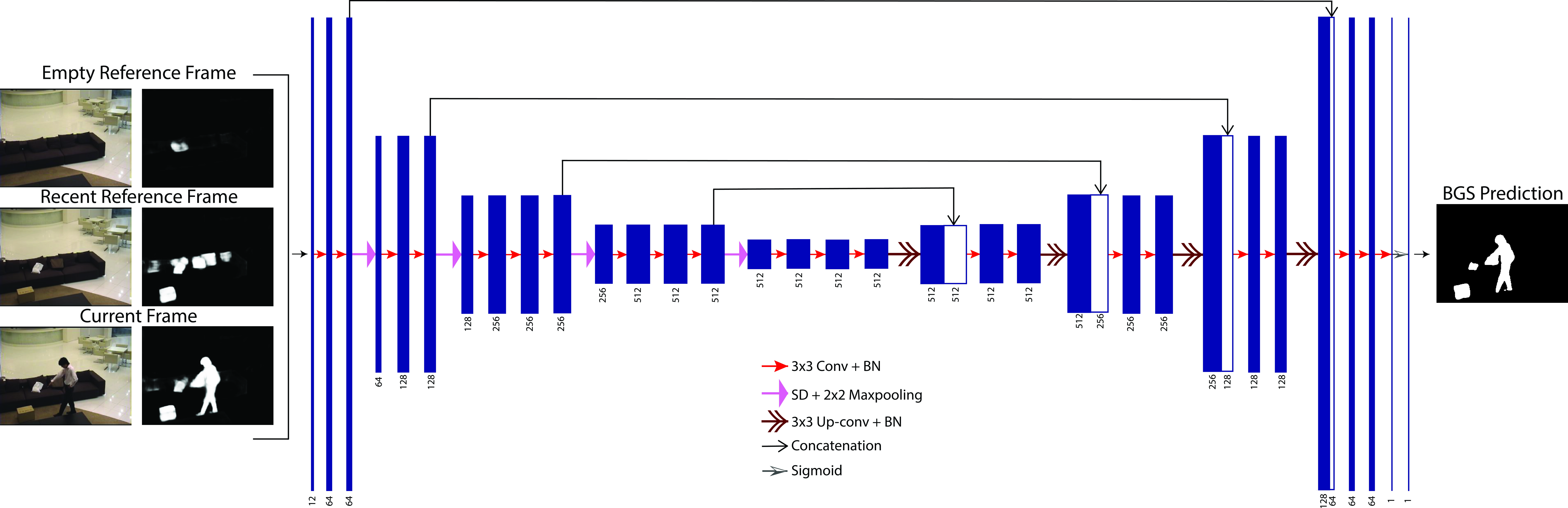}
  \caption{Network architecture of {\ourmodel}. BN stands for batch normalization and SD stands for spatial dropout. 
    Grayscale images at the network input show foreground probability maps (FPM) of the corresponding RGB frames.}
  \label{fig:network}
\end{figure*}

We use a UNET-type \cite{ronneberger2015unet} fully-convolutional neural network (FCNN) with residual connections. The architecture of {\ourmodel} has two parts: encoder and decoder, and is shown in Figure \ref{fig:network}. In the encoder network, we use $2 \times 2$ max-pooling operators to decrease the spatial dimensions. In the decoder network, we use up-convolutional layers (transposed convolution with a stride of 2) to increase the dimensions back to those of the input. In all convolutional and up-convolutional layers, we use $3 \times 3$ convolutions as in VGG \cite{simonyan2014vgg}. The residual connections from the encoder to the decoder help the network combine low-level visual information gained in the initial layers with high-level visual information gained in the deeper layers. Since our aim is to increase the performance on unseen videos, we use strong batch normalization (BN) \cite{ioffe2015batchnorm} and spatial dropout (SD) \cite{tompson2015spatialdropout} layers to increase the generalization capacity. Specifically, we use a BN layer after each convolutional and up-convolutional layer, and an SD layer before each max-pooling layer. Since our task can be viewed as a binary segmentation,
we use a sigmoid layer as the last layer in {\ourmodel}. The operation of the overall network can be defined as a nonlinear map $\mathbf{G}(\mathbf{W}): \mathbf{X} \rightarrow \widehat{\mathbf{Y}}$
where $\mathbf{X} \in \mathbb{R}^{w \times h \times 12}$ is a 12-channel input, $w$ and $h$ are its spatial dimensions, $\mathbf{W}$ represents the parameters of neural network $\mathbf{G}$, and $\widehat{\mathbf{Y}} \in [0,1]^{w \times h}$
is a pixel-wise foreground probability prediction. Note that since this is a fully-convolutional neural network, it does not require fixed input size; any frame size can be used, but some padding may be needed to account for max-pooling operations.
    
In most BGS datasets, the number of background pixels is much larger than the number of foreground pixels. This class imbalance creates significant problems for the commonly-used loss functions, such as cross-entropy and mean-squared error. A good alternative for unbalanced binary datasets is the Jaccard index. Since the network output is a probability map, 
we opted for a relaxed form of the Jaccard index as the loss function, defined as follows:
{\small
\begin{equation*}
J_R(\mathbf{Y}, \widehat{\mathbf{Y}}) \!=\!
\frac{T + \sum\limits_{m,n}(\mathbf{Y}[m,n] \widehat{\mathbf{Y}}[m,n])}{T \!+\!  \sum\limits_{m,n}\!\!\Big(\mathbf{Y}[m,n] \!+\! \widehat{\mathbf{Y}}[{m,n}] \!-\! \mathbf{Y}[{m,n}]  \widehat{\mathbf{Y}}[{m,n}] \Big)
    }
\end{equation*}}where $\mathbf{Y} \in \{0, 1\}^{w \times h}$ is the ground truth of $\mathbf{X}$, $T$ is a smoothing parameter and $m$, $n$ are the spatial locations.
    
\subsection{Resilience to Illumination Change by Data Augmentation}
\label{sec:illres}
Since neural networks have millions of parameters, they are very prone to overfitting. A widely-used method for reducing overfitting in computer-vision problems is to enlarge the dataset by applying several data augmentations such as random crops, rotations and noise addition. Since we are dealing with videos in this paper, we can also add augmentation in the temporal domain. 
    
In real-life BGS problems, there might be a significant illumination difference between an empty background frame acquired at an earlier time and the current frame. However, only a small portion of videos in CDnet-2014 capture significant illumination changes which limits \ourmodel's generalization performance.
Therefore, we introduce a new data-augmentation technique to account for global illumination changes between the empty reference frame and the current frame. Suppose that $\mathbf{R_E} \in \mathbb{R}^{w, h, 3}$ represents the RGB channels of an empty reference frame. Then, an augmented version of $\mathbf{R_E}$ can be computed as $\widehat{\mathbf{R}}_{\mathbf E}[m, n, c] = \mathbf{R_E}[m, n, c] + \mathbf{d}[c]$ for $c = 1,2,3$, where $\mathbf{d} \in \mathbb{R}^3$ represents a frame-specific global RGB change in our illumination model.
By choosing $\mathbf{d}$ randomly for each example during training (see Section~\ref{sec:train-evaluate} for details), we can make the network resilient to illumination variations.

\section{Experimental Results}
\label{sec:exp_res}

\subsection{Dataset and Evaluation Metrics}\label{sec:data_and_eval}

In order to evaluate the performance of BSUV-Net, we used CDNet-2014 \cite{goyette2012changedetection}, the largest BGS dataset with 53 natural videos from 11 categories including challenging scenarios such as shadows, night videos, dynamic background, etc. The spatial resolution of videos varies from $320 \times 240$ to $720 \times 526$ pixels. Each video has a region of interest labelled as either 1) foreground, 2) background, 3) hard shadow or 4) unknown motion. When measuring an algorithm's performance, we ignored pixels with unknown motion label and considered hard-shadow pixels as background. Our treatment of hard shadows is consistent with what is done in CDNet-2014 for the change-detection task.

\new{In CDNet-2014 \cite{goyette2012changedetection}, the authors propose seven binary performance metrics to cover a wide range of BGS cases: recall ($Re$), specificity ($Sp$), false positive rate ($FPR$), false negative rate ($FNR$), percentage of wrong classification ($PWC$), precision ($Pr$) and F-measure ($F_1$).
They also introduced two ranking-based metrics namely ``average ranking'' ($R$) and ``average ranking accross categories'' ($R_{cat}$) which combine all 7 metrics into ranking scores. The details of these rankings can be found at ``changedetection.net''. }

\subsection{Training and Evaluation Details}\label{sec:train-evaluate}
    
As discussed in Section \ref{sec:supervisedbgs}, we use a \textit{video-agnostic} evaluation methodology in all experiments. This allows us to measure an algorithm's performance on real-world-like tasks when no ground-truth labels are available. 
To evaluate {\ourmodel} performance on all videos in CDNet-2014, we used 18 different combinations of training/test video sets. 
\new{The splits are structured in such a manner that every video appears in the test set of exactly one split, but when it does so, it does not appear in the training set for that split. Detailed information about these sets can be found in the supplementary material.}
Let us denote the $m$-th combination by $(V_{train}^m, \ V_{test}^m)$. Then, $\cup_{m=1}^{18} V_{test}^m$ is equal to the set of all 53 videos in CDNet-2014. During training, we used 200 frames suggested in \cite{zeng2018mfcnnbgs} for each video in $V_{train}^m$. 
    
When training on different sets $V_{train}^m$, we used {\it exactly the same} hyperparameter values across all sets to make sure that we are not tuning our network to specific videos. In all of our experiments, we used ADAM optimizer with a learning rate of $10^{-4}$, $\beta_1 = 0.9$, and $\beta_2=0.99$. The minibatch size was 8 and we trained for 50 epochs.
As the empty background frame, we used the median of all foreground-free frames within the first 100 frames. In a few videos containing highway traffic, the first 100 frames did not contain a single foreground-free frame. For these videos, we hand-picked empty frames (e.g., in groups) and used their median as the empty reference. Although this may seem like a limitation, in practice one can randomly sample several hundred frames at the same time of the day across several days (similar illumination) and median filter them to obtain an empty background frame (due to random selection, a moving object is unlikely to occupy the same location in more than 50\% of frames). Since there is no single empty background frame in videos from the pan-tilt-zoom (PTZ) category, we slightly changed the inputs. Instead of ``empty background + recent background'' pair we used ``recent background + more recent background'' pair, where the recent background is computed as the median of 100 preceding frames and the more recent background is computed as the median of 30 preceding frames.
    
Although {\ourmodel} can accept frames of any spatial dimension, we used a {\it fixed} size of $224 \times 224$ pixels (randomly cropped from the input frame) so as to leverage parallel GPU processing in the training process. We applied random data augmentation at the beginning of each epoch. For illumination resilience, we used the data augmentation method of Section~\ref{sec:illres} with $\mathbf{d}[c] = \mathbf{I} + \mathbf{I}_c$, where $\mathbf{I}$ is the same for all $c$ and sampled from $\mathcal{N}(0, 0.1^2)$, while $\mathbf{I}_c$ is independently sampled from  $\mathcal{N}(0, 0.04^2)$ for each $c$.
We also added random Gaussian noise from  $\mathcal{N}(0, 0.01^2)$ to each pixel in each color channel.
For pixel values, we used double precision numbers that lie between $0$ and $1$.
    
In the evaluation step, we did not apply any scaling or cropping to the inputs. To obtain binary maps, we applied thresholding with  threshold $\theta=0.5$ to the output of the sigmoid layer of {\ourmodel}. 
    
\subsection{Quantitative Results}
\label{sec:results}
  
\begin{table*}[t]
  \centering
  \caption{
  \new{Comparison of methods for {\bf unseen videos} from CDNet-2014. For fairness, we separated the post-processing and self-contained algorithms.}
  }
  \smallskip
        \begin{tabular}{c|cc|ccccccc}
            \hline
             Method & $R$ & $R_{cat}$ & $Re$ & $Sp$ & $FPR$ & $FNR$ & $PWC$ & $Pr$ & $F_1$\\
             \hline
             \multicolumn{10}{c}{\textit{Post-processing algorithms}} \\
             {\bf BSUV-net} + SemanticBGS$^*$  & \textbf{9.00} & 14.00 & \textbf{0.8179} & 0.9944 & 0.0056 & \textbf{0.1821} & 1.1326 & \textbf{0.8319} & \textbf{0.7986}\\
             IUTIS-5$^*$ + SemanticBGS$^*$ & 9.43 & 11.45 & 0.7890 &	\textbf{0.9961} &	\textbf{0.0039} &	0.2110 &	\textbf{1.0722} &	0.8305 & 0.7892\\
             IUTIS-5$^*$ & 11.43 & \textbf{10.36} & 0.7849 &	0.9948 &	0.0052 &	0.2151 &	1.1986 &	0.8087 & 0.7717	\\
             \multicolumn{10}{c}{\textit{\Basealgo{}}} \\
             {\bf BSUV-net}  & \textbf{9.29} & \textbf{13.18} & \textbf{0.8203} &	0.9946 &	0.0054 &	\textbf{0.1797} &	\textbf{1.1402} &	\textbf{0.8113} &	\textbf{0.7868}\\
             SWCD & 15.43 & 19.00 & 0.7839 &	0.9930 &	0.0070 &	0.2161 &	1.3414 &	0.7527 &	0.7583\\
             WisenetMD & 16.29 & 15.18 & 0.8179 &	0.9904 &	0.0096 &	0.1821 &	1.6136 &	0.7535 &	0.7668\\ 
             PAWCS & 14.00 & 15.45 & 0.7718 &	\textbf{0.9949} &	\textbf{0.0051} &	0.2282 &	1.1992 &	0.7857 &	0.7403\\
             FgSegNet v2 &	44.57 &44.09&	0.5119&	0.9411&	0.0589	&0.4881&	7.3507&	0.4859&	0.3715\\
             \hline
        \end{tabular}

  \label{table:overall_results}
\end{table*}
\begin{table*}[t]
  \centering
  \caption{
  \new{Comparison of methods according to the per-category F-measure for {\bf unseen videos} from CDNet-2014.}
  }
  \smallskip
  \scalebox{0.8}{
        \begin{tabular}{c|ccccccccccc|c}
            \hline
             Method & \!\!\! \makecell{Bad\\weather}\! & \!\! \makecell{Low\\framerate}\!\!\! & Night & \!\!\! PTZ & \!\!\!\!\!Thermal\!\! & \!\!\!\! Shadow\!\!&   \!\! \makecell{Int.\ obj.\\motion} &\!\!\!\! \makecell{Camera\\jitter} \! & \!\! \makecell{Dynamic\\backgr.}\!\!  & \makecell{Base-\\line} & \makecell{Turbu-\\lence} & Overall\\
             \hline
             \multicolumn{13}{c}{\textit{Post-processing algorithms}} \\
             \!\!{\bf BSUV-net} + SemanticBGS$^*$ & \textbf{0.8730} & 0.6788 & \textbf{0.6815} & \textbf{0.6562} & \textbf{0.8455} & \textbf{0.9664} & 0.7601 & 0.7788 & 0.8176 & \textbf{0.9640} & 0.7631 & \textbf{0.7986}\\
             IUTIS-5$^*$ + SemanticBGS$^*$ & 0.8260 & 0.7888 & 0.5014 & 0.5673 & 0.8219 & 0.9478 & \textbf{0.7878} & \textbf{0.8388} & \textbf{0.9489} & 0.9604 & 0.6921 & 0.7892\\
             IUTIS-5$^*$ & 0.8248 & \textbf{0.7743} & 0.5290 & 0.4282 & 0.8303 & 0.9084 & 0.7296 
             & 0.8332 & 0.8902 & 0.9567 & \textbf{0.7836} & 0.7717\\
             \multicolumn{13}{c}{\textit{\Basealgo{}}} \\
             {\bf BSUV-net}  & \textbf{0.8713}&	0.6797&	\textbf{0.6987}&	\textbf{0.6282}&	\textbf{0.8581}&	0.9233&	0.7499&	0.7743&	0.7967&	\textbf{0.9693}&	0.7051	& 0.7868\\
             RTSS & 0.8662 & 0.6771 & 0.5295 & 0.5489 & 0.8510 & \textbf{0.9551} & \textbf{0.7864} & \textbf{0.8396} & \textbf{0.9325} & 0.9597 & 0.7630 & \textbf{0.7917}\\
             SWCD & 0.8233 & \textbf{0.7374} & 0.5807 & 0.4545 & \textbf{0.8581}  & 0.8779 & 0.7092 
             & 0.7411  & 0.8645 & 0.9214 & 0.7735 & 0.7583\\
             WisenetMD & 0.8616 & 0.6404 & 0.5701 & 0.3367 &  0.8152 & 0.8984 & 0.7264 
             & 0.8228  & 0.8376 & 0.9487 & \textbf{0.8304}& 0.7535\\
             PAWCS  & 0.8152 & 0.6588 & 0.4152 & 0.4615 & 0.8324 & 0.8913 & 0.7764 
             & 0.8137 & 0.8938 & 0.9397 & 0.6450 & 0.7403\\
             FgSegNet v2 & 0.3277 & 0.2482 & 0.2800 & 0.3503 & 0.6038 & 0.5295 & 0.2002 & 0.4266 & 0.3634 & 0.6926 & 0.0643 & 0.3715\\
             \hline
        \end{tabular}
  }
  \label{table:cat_results}
\end{table*}
    
\new{Table~\ref{table:overall_results} compares {\ourmodel} against state-of-the-art BGS algorithms in terms of the seven  metrics and two rankings discussed in Section \ref{sec:data_and_eval}. All quantitative results shown in this paper are computed by ``changedetection.net'' evaluation servers to reflect the real performance on test data. Since {\ourmodel} is \textit{video-agnostic}, comparing it with \textit{video-optimized} or \textit{video-group-optimized} algorithms would not be fair and we omit them. Instead, we compare BSUV-Net with state-of-the-art unsupervised algorithms, namely  SWCD \cite{icsik2018swcd}, WisenetMD \cite{lee2018wisenetmd} and PAWCS \cite{st2015pawcs} 
, which, by definition, are {\it video-agnostic}. We exclude RTSS \cite{zeng2019rtss} and $3DFR$ \cite{mandal20193dfr} in Table~\ref{table:overall_results} since their results on the test frames are not available on ``changedetection.net''.
We include the results of IUTIS-5 \cite{bianco2017iutis} and SemanticBGS \cite{braham2017semanticbgs}, but we list them separately because these are \textit{post-processing} algorithms. Note that, both IUTIS-5 and SemanticBGS can be applied to any BGS algorithm from Table~\ref{table:overall_results}, including {\ourmodel}, to improve its performance. To show this, we also report the result of {\ourmodel} post-processed by SemanticBGS. In the \textit{\basealgo{}} category, we also list FgSegNet v2 \cite{lim2018fgsegnetv3} since it is currently the best performing algorithm on CDNet-2014.
However, since FGSegNet v2's performance reported on ``changedetection.net'' has been obtained in a \textit{video-optimized} manner, we trained it anew in a {\it video-agnostic} manner using the same methodology that we used for BSUV-Net.
As expected, this caused a huge performance decrease of FgSegNet v2 compared to it’s \textit{video-optimized} training. As is clear from Table~\ref{table:overall_results}, {\ourmodel} outperforms its competitors on almost all of the metrics. The F-measure performance demonstrates that {\ourmodel} achieves excellent results without compromising either recall or precision. Table~\ref{table:overall_results} also shows that the performance of {\ourmodel} can be improved even further by combining it with SemanticBGS. The combined algorithm outperforms all of the {\it video-agnostic} algorithms that are available on ``changedetection.net''.}
    
Table~\ref{table:cat_results} compares the per-category F-measure performance of {\ourmodel} against state-of-the-art BGS algorithms.For RTSS \cite{zeng2019rtss}, the values of performance metrics shown in Table~\ref{table:cat_results} are as reported in their paper.

Columns 2-12 report the F-measure for each of the 11 categories from ``changedetection.net'', while the last column reports the mean F-measure across all categories. Similarly to Table~\ref{table:overall_results}, we divided this table into \textit{post-processing} and \textit{\basealgo{}}. 
It can be observed that {\ourmodel} achieves the best performance in 5 out of 11 categories. It has a striking performance advantage in the ``night'' category. All videos in this category are traffic-related and many cars have headlights turned on at night which causes significant local illumination variations in time. {\ourmodel}'s excellent performance in this category demonstrates that the proposed model is indeed largely illumination-invariant.
    
\new{{\ourmodel} performs poorer than other algorithms in ``camera jitter'' and ``dynamic background'' categories. We believe this is related to the empty and recent background frames we are using as input. The median operation used to compute background frames creates very blurry images for these categories since the background is not static. Thus, {\ourmodel} predicts some pixels in the background as foreground and increases the number of false positives.}
     
\subsection{Visual Results} 

\begin{figure*}
  \centering
  \includegraphics[width=\textwidth]{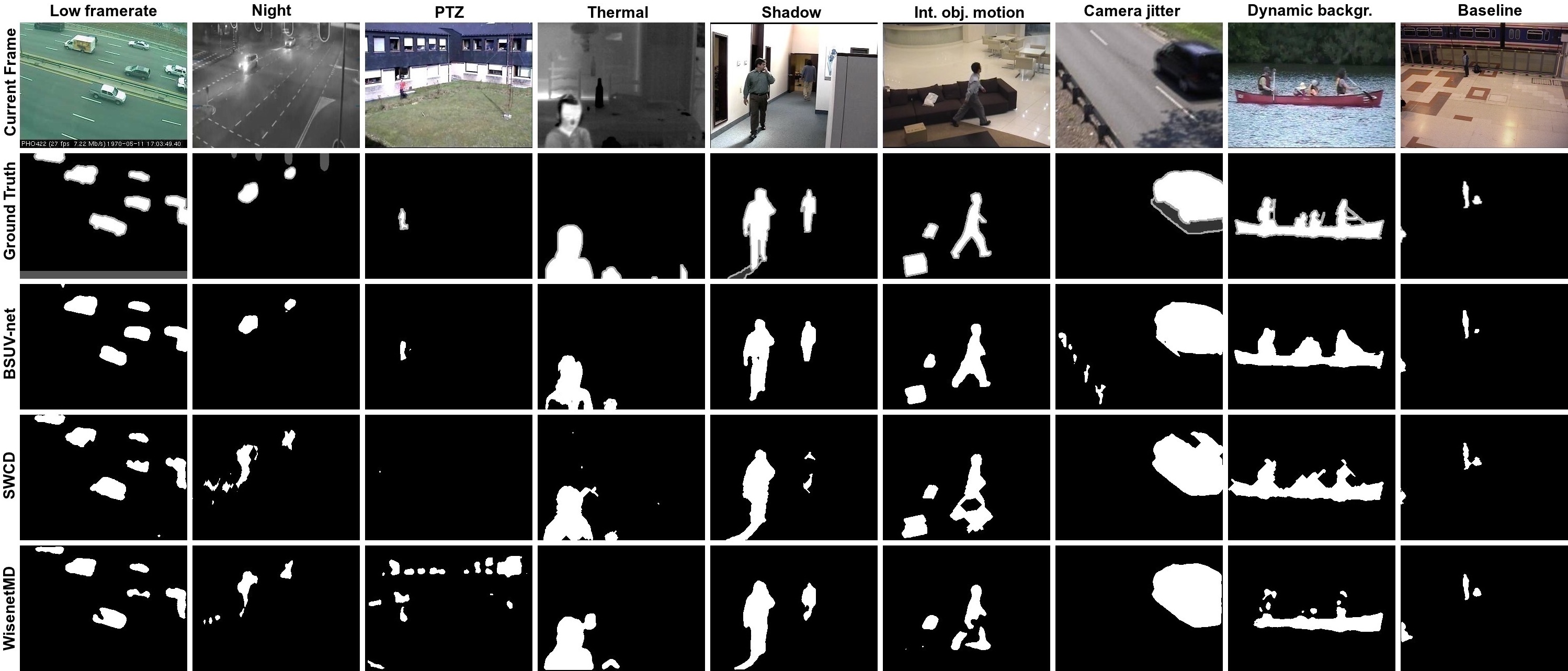}
  \caption{
  \new{Visual comparison of sample results from {\ourmodel}, SWCD and WisenetMD on {\bf unseen videos} from CDNet-2014.}
  }
  \label{fig:vis_results}
\end{figure*}

A visual comparison of {\ourmodel} with SWCD \cite{icsik2018swcd} and WisenetMD \cite{lee2018wisenetmd} is shown in Figure \ref{fig:vis_results}. Each column shows a sample frame from one of the videos in one of the 8 categories.
It can be observed that {\ourmodel} produces visually the best results for almost all categories.

In the ``night'' category, SWCD and WisenetMD produce many false positives because of local illumination changes. {\ourmodel} produces better results since it is designed to be illumination-invariant. In the ``shadow'' category, {\ourmodel} performs much better in the shadow regions.
Results in the ``intermittent object motion'' and ``baseline'' categories show that {\ourmodel} can successfully detect intermittently-static objects. It is safe to say that {\ourmodel} is capable of simultaneously handling the discovery of intermittently-static objects and also the dynamic factors such as illumination change.

An inspection of results in the ``dynamic background'' category shows that {\ourmodel} has detected most of the foreground pixels but failed to detect the background pixels around the foreground objects. We believe this is due to the blurring effect of the median operation that we used in the computation of background frames. Using more advanced background models as an input to {\ourmodel} might improve the performance in this category.

\subsection{Ablation Study}

\new{One of the contributions of {\ourmodel} is its multi-channel input composed of two background frames from different time scales and a foreground probability map (FPM). Another contribution is temporal data augmentation tailored to handling illumination changes. In Table~\ref{table:ablation}, we explore their impact on precision, recall and F-measure. Each column on the left represents one characteristic and each row represents a different combination of these characteristics. RGB channels of the current frame are used in all of the combinations. ``Empty BG'' and ``Recent BG'' refer to the use of empty and$\backslash$or recent background frames, respectively, in addition to the current frame. ``Data aug.'' refers to temporal data augmentation described in Section~\ref{sec:illres} and ``FPM'' refers to the use of semantic FPM channel in addition to the RGB channels for all input frames. It is clear that all these characteristics have a significant impact on the overall performance. Using only the current frame as input results in very poor metrics. The introduction of empty or/and recent background frames leads to a significant improvement. Adding temporal data augmentation or/and FPM channels further improves the performance and the final network achieves state-of-the-art results.}
\begin{table}[ht]
    \centering
    \caption{
    \new{Impact of background frames, data augmentation for temporal illumination change and FPM on {\ourmodel} performance.}
    }
    \smallskip
    \new{\scalebox{0.85}{
        \begin{tabular}{ccccc|ccc}
            \hline
             \!\!\makecell{Current \\ frame}\!\! & \!\!\makecell{Empty \\ BG}\!\! & \!\!\makecell{Recent \\ BG}\!\! &\!\!\makecell{Data \\ aug.}\!\!& \!\!FPM\!\!& $Pr$\!& \!$Re$\!& \!$F_1$\\
             \hline
             \hline
              \cmark && & & &  0.3615 & 0.5509 & 0.3476\\
              \hline
             \cmark &\cmark & & & & 0.6994 & 0.7686 & 0.6819\\
             \hline
             \cmark & & \cmark & & &0.6976 & 0.7064 & 0.6351\\
              \hline
             \cmark &\cmark & \cmark & & & 0.7658 & 0.7606 & 0.7156\\
             \hline
             \cmark &\cmark &\cmark &\cmark & & 0.7574 & 0.8159 & 0.7447\\
             \hline
             \cmark &\cmark &\cmark & &\cmark &  0.7807 & 0.7747 & 0.7450\\
             \hline
             \cmark &\cmark &\cmark &\cmark &\cmark & 0.8113 & 0.8203 & 0.7868\\
             \hline
        \end{tabular}}
        }
        \label{table:ablation}
\end{table}

In this paper, we proposed to add semantic FPM channel as input in order to improve our algorithm's performance. However, if the background and foreground object categories are chosen carefully, FPM can be used as a BGS algorithm by itself. This would require prior information about the video (to compute FPM) and, therefore, would not qualify as a \textit{video-agnostic} method. In our algorithm, however, we combine FPM information with RGB channels and background frames. When applying DeepLabv3 \cite{braham2017semanticbgs} to compute FPM frames, we pre-defined global background and foreground class categories which might be wrong for some of the videos. We did not optimize the selection of these class categories but instead used those suggested in \cite{braham2017semanticbgs}. To demonstrate that our algorithm is not replicating FPM but leverages its semantic information to boost performance, we compared {\ourmodel} with thresholded FPM used as a BGS result (Table~\ref{table:FPM}). It is clear that FPM alone is not a powerful tool for BGS as it is significantly outperformed by {\ourmodel}.

\begin{table}[ht]
    \centering
    \caption{
    \new{Comparison of {\ourmodel} with thresholded FPM used as a BGS result (probability threshold equals 0.5).}
    }
    \smallskip
        \begin{tabular}{c|ccc}
             Method & $Pr$ & $Re$& $F_1$\\
             \hline \hline
             FPM & 0.6549 & 0.6654 & 0.5846 \\
             BSUV-net & 0.8113 & 0.8203 & 0.7868\\
        \end{tabular}
        \label{table:FPM}
\end{table}

While in {\ourmodel} we assume that the \textit{empty background frame} is foreground-free, CDNet-2014 does not provide empty background frames. Therefore, in some videos, we manually selected empty background frames from among the initial frames as explained in Section~\ref{sec:train-evaluate}. In Table~\ref{table:empty}, we show the impact of this manual process by comparing the manual selection strategy with an automatic one, that is using the median of all frames in the test video as a the \textit{empty background frame}.
Clearly, the manual selection slightly improves precision while significantly decreasing recall. We believe this is due to the increase of false negatives caused by the appearance of some of the foreground objects in the empty background. Since videos in CDNet2014 are rather short (at most 10 minutes), in some cases the median of all frames does not result in an empty background. However, for stationary surveillance cameras in a real-life scenario it is often possible to compute an empty background, for example by taking the median of frames at the same time of the day (when it is expected to be empty) over many days.

\begin{table}[ht]
    \centering
    \caption{
    \new{Comparison of manual and automatic selection of empty background frames in {\ourmodel}.}
    }
    \smallskip
        \begin{tabular}{c|ccc}
             \makecell{Empty background \\ selection} & $Pr$& $Re$& $F_1$\\
             \hline \hline
             Automatic & 0.8207 & 0.7812 & 0.7639\\
             Manual & 0.8113 & 0.8203 & 0.7868\\
        \end{tabular}
        \label{table:empty}
\end{table}

\section{Conclusions and Future Work}
\label{sec:discuss}
We introduced a novel deep-learning algorithm for background subtraction of unseen videos and proposed a \textit{video-agnostic} evaluation methodology that treats each video in a dataset as unseen. The input to {\ourmodel} consists of the current frame and two reference frames from different time-scales, along with semantic information for all three frames (computed using Deeplabv3 \cite{chen2017deeplab}). To increase the generalization capacity of {\ourmodel}, we formulated a simple, yet effective, illumination-change model. 
Experimental results on CDNet-2014 show that {\ourmodel} outperforms state-of-the-art \textit{video-agnostic} BGS algorithms in terms of 7 out of 9 performance metrics. Its performance can be further improved by adding SemanticBGS \cite{braham2017semanticbgs} as a post-processing layer. This shows great potential for deep-learning BGS algorithms designed for unseen or unlabeled videos.

In the future, we are planning further work on temporal data-augmentation techniques to improve performance for challenging categories, such as ``dynamic background'' and ``camera jitter''. We will also investigate different background models for the reference frames. In this work, we kept our focus on designing a high-performance, supervised BGS algorithm for unseen videos without considering the computational complexity. To bring {\ourmodel} closer to real-time performance, we are also planning to study a shallow-network implementation designed for fast inference.

\bibliographystyle{plain}
\bibliography{strings,ref}

\clearpage

\thispagestyle{empty}

\twocolumn[
\section*{\centering{\huge{Appendices}}}]

\section*{Selection of training test sets for \\ evaluation on unseen videos}
In this paper, we introduced a supervised background subtraction (BGS) algorithm for unseen videos. As for all supervised learning algorithms, the size and diversity of the training data are crucially important for the learning process. Generally speaking, for most of the state-of-the-art deep neural networks, the best approach is to use all of the available training data. Unfortunately, CDNet 2014 \cite{goyette2012changedetection} does not provide different videos for training and testing. Instead, it provides some frames from each video as training data and the remaining ones -- as test data. However, this type of division is not useful for evaluating the performance on {\it unseen} videos. 

For comparing the performance of different models on \textit{unseen} videos, we split the dataset into 18 different sets of training/testing videos as shown in Tables~\ref{table:traintest1} and \ref{table:traintest2}. When training a supervised algorithm, the main assumption is that the training set is diverse enough to cover a wide range of test scenarios. For example, if there are no examples that include shadow in the training set, then it is impossible for the network to learn how to classify shadow regions. Therefore, we designed the splits so that the training set for each split contains some videos from the same category as the test videos. 
We did not perform a full ``leave-$k$-videos-out'' cross-validation due to prohibitive time need to train {\ourmodel}. 
In all of the tests, we used videos from ``baseline'', ``bad weather'', ``intermittent object motion'', ``low frame rate'' and ``shadow'' categories during training since they span most of the common scenarios. 
For videos from more difficult scenarios, we progressively added additional categories into the training set.
In particular, we considered ``PTZ'', ``thermal'' and ``turbulence'' categories as the most difficult ones since they have substantially different data characteristics from other categories. ``PTZ'' is the only category with significant camera movement and zoom in/out, while ``thermal'' and `turbulence'' categories capture different scene properties than the remaining categories (far- and near-infrared spectrum instead of RGB, respectively). For these 3 categories, we used more videos in the training set, than in the other categories. 
Please note that a '`leave-$k$-videos-out'' approach would have more videos in the training set compared to our splits and is therefore likely to yield better results.

\begin{table*}
\floatpagestyle{empty}
\centering
\caption{Training and test splits $\mathbf{S_1}$ to $\mathbf{S_{12}}$ used for evaluation.}
\scalebox{0.92}{
\begin{tabular}{|c|c|c|c|c|c|c|c|c|c|c|c|c|c|}
    \hline
     \textbf{category} & \textbf{video}  & $\mathbf{S_1}$& $\mathbf{S_2}$& $\mathbf{S_3}$& $\mathbf{S_4}$& $\mathbf{S_5}$& $\mathbf{S_6}$& $\mathbf{S_7}$& $\mathbf{S_8}$& $\mathbf{S_9}$& $\mathbf{S_{10}}$& $\mathbf{S_{11}}$& $\mathbf{S_{12}}$ \\

    \hline \hline 
    \multirow{4}{*}{baseline} & highway & \color{blue}{Tr} & \color{red}{Test} & \color{blue}{Tr} & \color{blue}{Tr} & \color{blue}{Tr} & \color{blue}{Tr} & \color{blue}{Tr} & \color{blue}{Tr} & \color{blue}{Tr} & \color{blue}{Tr}& \color{blue}{Tr} & \color{blue}{Tr}\\ 
    \cline{2-14}
     & pedestrians & \color{blue}{Tr} & \color{red}{Test} & \color{blue}{Tr} & \color{blue}{Tr} & \color{blue}{Tr} & \color{blue}{Tr} & \color{blue}{Tr} & \color{blue}{Tr}& \color{blue}{Tr} & \color{blue}{Tr} & \color{blue}{Tr} & \color{blue}{Tr}\\ 
    \cline{2-14}
     & office & \color{red}{Test} & \color{blue}{Tr} & \color{blue}{Tr} & \color{blue}{Tr} & \color{blue}{Tr} & \color{blue}{Tr} & \color{blue}{Tr} & \color{blue}{Tr} & \color{blue}{Tr} & \color{blue}{Tr}& \color{blue}{Tr} & \color{blue}{Tr}\\ 
    \cline{2-14}
     & PETS2006 & \color{red}{Test} & \color{blue}{Tr} & \color{blue}{Tr} & \color{blue}{Tr} & \color{blue}{Tr} & \color{blue}{Tr} & \color{blue}{Tr} & \color{blue}{Tr} & \color{blue}{Tr} & \color{blue}{Tr}& \color{blue}{Tr} & \color{blue}{Tr}\\

    \hline \hline 
    \multirow{4}{*}{\makecell{bad\\weather}} & skating & \color{blue}{Tr} & \color{blue}{Tr} & \color{red}{Test} & \color{blue}{Tr} & \color{blue}{Tr} & \color{blue}{Tr} &  \color{blue}{Tr} & \color{blue}{Tr}  & \color{blue}{Tr} & \color{blue}{Tr}& \color{blue}{Tr} & \color{blue}{Tr}\\ 
    \cline{2-14}
     & blizzard & \color{blue}{Tr} & \color{blue}{Tr} & \color{red}{Test} & \color{blue}{Tr} & \color{blue}{Tr} & \color{blue}{Tr} &  \color{blue}{Tr} & \color{blue}{Tr}  & \color{blue}{Tr} & \color{blue}{Tr}& \color{blue}{Tr} & \color{blue}{Tr}\\ 
    \cline{2-14}
     & snowFall & \color{blue}{Tr} & \color{blue}{Tr} & \color{blue}{Tr} & \color{red}{Test} & \color{blue}{Tr} & \color{blue}{Tr} &  \color{blue}{Tr} & \color{blue}{Tr}  & \color{blue}{Tr} & \color{blue}{Tr}& \color{blue}{Tr} & \color{blue}{Tr}\\ 
    \cline{2-14}
     & wetSnow & \color{blue}{Tr} & \color{blue}{Tr} & \color{blue}{Tr} & \color{red}{Test} & \color{blue}{Tr} & \color{blue}{Tr} &  \color{blue}{Tr} & \color{blue}{Tr}  & \color{blue}{Tr} & \color{blue}{Tr}& \color{blue}{Tr} & \color{blue}{Tr}\\

    \hline \hline 
    \multirow{6}{*}{\makecell{intermittent\\object\\motion}} & abandonedBox & \color{blue}{Tr} & \color{blue}{Tr} & \color{blue}{Tr} & \color{blue}{Tr} & \color{blue}{Tr} & \color{red}{Test} & \color{blue}{Tr} & \color{blue}{Tr} & \color{blue}{Tr} & \color{blue}{Tr}& \color{blue}{Tr} & \color{blue}{Tr}\\ 
    \cline{2-14}
     & parking & \color{blue}{Tr} & \color{blue}{Tr} & \color{blue}{Tr} & \color{blue}{Tr} & \color{red}{Test} & \color{blue}{Tr} & \color{blue}{Tr} & \color{blue}{Tr} & \color{blue}{Tr} & \color{blue}{Tr}& \color{blue}{Tr} & \color{blue}{Tr}\\ 
    \cline{2-14}
     & sofa & \color{blue}{Tr} & \color{blue}{Tr} & \color{blue}{Tr} & \color{blue}{Tr} & \color{blue}{Tr} & \color{red}{Test} & \color{blue}{Tr} & \color{blue}{Tr} & \color{blue}{Tr} & \color{blue}{Tr}& \color{blue}{Tr} & \color{blue}{Tr}\\ 
    \cline{2-14}
     & streetLight & \color{blue}{Tr} & \color{blue}{Tr} & \color{blue}{Tr} & \color{blue}{Tr} & \color{blue}{Tr} & \color{red}{Test} & \color{blue}{Tr} & \color{blue}{Tr} & \color{blue}{Tr} & \color{blue}{Tr}& \color{blue}{Tr} & \color{blue}{Tr}\\ 
    \cline{2-14}
     & tramstop & \color{blue}{Tr} & \color{blue}{Tr} & \color{blue}{Tr} & \color{blue}{Tr} & \color{red}{Test} & \color{blue}{Tr} & \color{blue}{Tr} & \color{blue}{Tr} & \color{blue}{Tr} & \color{blue}{Tr}& \color{blue}{Tr} & \color{blue}{Tr}\\ 
    \cline{2-14}
     & winterDriveway & \color{blue}{Tr} & \color{blue}{Tr} & \color{blue}{Tr} & \color{blue}{Tr} & \color{red}{Test} & \color{blue}{Tr} & \color{blue}{Tr} & \color{blue}{Tr} & \color{blue}{Tr} & \color{blue}{Tr}& \color{blue}{Tr} & \color{blue}{Tr}\\

    \hline \hline 
    \multirow{4}{*}{\makecell{low\\framerate}} & port 0.17fps & \color{blue}{Tr} & \color{blue}{Tr} & \color{red}{Test} & \color{blue}{Tr} & \color{blue}{Tr} & \color{blue}{Tr} &  \color{blue}{Tr} & \color{blue}{Tr}  & \color{blue}{Tr} & \color{blue}{Tr}& \color{blue}{Tr} & \color{blue}{Tr}\\ 
    \cline{2-14}
     & tramCrossroad 1fps & \color{blue}{Tr} & \color{blue}{Tr} & \color{red}{Test} & \color{blue}{Tr} & \color{blue}{Tr} & \color{blue}{Tr} &  \color{blue}{Tr} & \color{blue}{Tr} & \color{blue}{Tr} & \color{blue}{Tr}  & \color{blue}{Tr} & \color{blue}{Tr}\\ 
    \cline{2-14}
     & tunnelExit 0.35fps & \color{blue}{Tr} & \color{blue}{Tr} & \color{blue}{Tr} & \color{red}{Test} & \color{blue}{Tr} & \color{blue}{Tr} & \color{blue}{Tr} & \color{blue}{Tr} & \color{blue}{Tr} & \color{blue}{Tr}& \color{blue}{Tr} & \color{blue}{Tr}\\ 
    \cline{2-14}
     & turnpike 0.5fps & \color{blue}{Tr} & \color{blue}{Tr} & \color{blue}{Tr} & \color{red}{Test} & \color{blue}{Tr} & \color{blue}{Tr} &  \color{blue}{Tr} & \color{blue}{Tr}  & \color{blue}{Tr} & \color{blue}{Tr}& \color{blue}{Tr} & \color{blue}{Tr}\\

    \hline \hline 
    \multirow{6}{*}{shadow} & backdoor & \color{red}{Test} & \color{blue}{Tr} & \color{blue}{Tr} & \color{blue}{Tr} & \color{blue}{Tr} & \color{blue}{Tr} & \color{blue}{Tr} & \color{blue}{Tr} & \color{blue}{Tr} & \color{blue}{Tr}& \color{blue}{Tr} & \color{blue}{Tr}\\ 
    \cline{2-14}
     & bungalows & \color{blue}{Tr} & \color{red}{Test} & \color{blue}{Tr} & \color{blue}{Tr} & \color{blue}{Tr} & \color{blue}{Tr} & \color{blue}{Tr} & \color{blue}{Tr} & \color{blue}{Tr} & \color{blue}{Tr}& \color{blue}{Tr} & \color{blue}{Tr}\\ 
    \cline{2-14}
     & busStation & \color{blue}{Tr} & \color{red}{Test} & \color{blue}{Tr} & \color{blue}{Tr} & \color{blue}{Tr} & \color{blue}{Tr} & \color{blue}{Tr} & \color{blue}{Tr} & \color{blue}{Tr} & \color{blue}{Tr}& \color{blue}{Tr} & \color{blue}{Tr}\\ 
    \cline{2-14}
     & copyMachine & \color{red}{Test} & \color{blue}{Tr} & \color{blue}{Tr} & \color{blue}{Tr} & \color{blue}{Tr} & \color{blue}{Tr} & \color{blue}{Tr} & \color{blue}{Tr} & \color{blue}{Tr} & \color{blue}{Tr}& \color{blue}{Tr} & \color{blue}{Tr}\\ 
    \cline{2-14}
     & cubicle & \color{blue}{Tr} & \color{red}{Test} & \color{blue}{Tr} & \color{blue}{Tr} & \color{blue}{Tr} & \color{blue}{Tr} & \color{blue}{Tr} & \color{blue}{Tr} & \color{blue}{Tr} & \color{blue}{Tr}& \color{blue}{Tr} & \color{blue}{Tr}\\ 
    \cline{2-14}
     & peopleInShade & \color{red}{Test} & \color{blue}{Tr} & \color{blue}{Tr} & \color{blue}{Tr} & \color{blue}{Tr} & \color{blue}{Tr} & \color{blue}{Tr} & \color{blue}{Tr} & \color{blue}{Tr} & \color{blue}{Tr}& \color{blue}{Tr} & \color{blue}{Tr}\\

    \hline \hline 
    \multirow{4}{*}{\makecell{camera\\jitter}} & badminton &  &  &  &  &  &  & \color{red}{Test} & \color{blue}{Tr} &  &  &  & \\ 
    \cline{2-14}
     & traffic &  &  &  &  &  &  & \color{red}{Test} & \color{blue}{Tr} &  &  &  & \\ 
    \cline{2-14}
     & boulevard &  &  &  &  &  &  & \color{blue}{Tr} & \color{red}{Test} &  &  &  & \\ 
    \cline{2-14}
     & sidewalk &  &  &  &  &  &  & \color{blue}{Tr} & \color{red}{Test} &  &  &  & \\

    \hline \hline 
    \multirow{6}{*}{\makecell{dynamic\\background}} & boats &  &  &  &  &  &  &  &  & \color{blue}{Tr} & \color{red}{Test}  &  &\\ 
    \cline{2-14}
     & canoe &  &  &  &  &  &  &  &  & \color{red}{Test} & \color{blue}{Tr}  &  &\\ 
    \cline{2-14}
     & fall &  &  &  &  &  &  &  &  & \color{red}{Test} & \color{blue}{Tr}  &  &\\ 
    \cline{2-14}
     & fountain01 &  &  &  &  &  &  &  &  & \color{blue}{Tr} & \color{red}{Test}  &  &\\ 
    \cline{2-14}
     & fountain02 &  &  &  &  &  &  &  &  & \color{red}{Test} & \color{blue}{Tr}  &  &\\ 
    \cline{2-14}
     & overpass &  &  &  &  &  &  &  &  & \color{blue}{Tr} & \color{red}{Test}  &  &\\

    \hline \hline 
    \multirow{6}{*}{\makecell{night\\videos}} & bridgeEntry & & &  &  & & &  &  &  &  & \color{red}{Test} & \color{blue}{Tr}\\ 
    \cline{2-14}
     & busyBoulvard & & &  &  & & &  &  &  &  & \color{blue}{Tr} & \color{red}{Test}\\ 
    \cline{2-14}
     & fluidHighway & & &  &  & & &  &  &  &  & \color{red}{Test} & \color{blue}{Tr}\\ 
    \cline{2-14}
     & streetCornerAtNight & & &  &  & & &  &  &  &  & \color{blue}{Tr} & \color{red}{Test}\\ 
    \cline{2-14}
     & tramStation & & &  &  & & &  &  &  &  & \color{red}{Test} & \color{blue}{Tr}\\ 
    \cline{2-14}
     & winterStreet & & &  &  & & &  &  &  &  & \color{blue}{Tr} & \color{red}{Test}\\

    \hline \hline 
    \multirow{4}{*}{PTZ} & continuousPan &  &  &  &  &  &  &  &  &  &  &  & \\ 
    \cline{2-14}
     & intermittentPan &  &  &  &  &  &  &  &  &  &  &  & \\ 
    \cline{2-14}
     & twoPositionPTZCam &  &  &  &  &  &  &  &  &  &  &  & \\ 
    \cline{2-14}
     & zoomInZoomOut &  &  &  &  &  &  &  &  &  &  &  & \\

    \hline \hline 
    \multirow{5}{*}{thermal} & corridor &  &  &  &  &  &  &  &  &  &  &  & \\ 
    \cline{2-14}
     & diningRoom &  &  &  &  &  &  &  &  &  &  &  & \\ 
    \cline{2-14}
     & lakeSide &  &  &  &  &  &  &  &  &  &  &  & \\ 
    \cline{2-14}
     & library &  &  &  &  &  &  &  &  &  &  &  & \\ 
    \cline{2-14}
     & park &  &  &  &  &  &  &  &  &  &  &  & \\

    \hline \hline 
    \multirow{4}{*}{turbulence} & turbulence0 &  &  &  &  &  &  &  &  &  &  &  & \\ 
    \cline{2-14}
     & turbulence1 &  &  &  &  &  &  &  &  &  &  &  & \\ 
    \cline{2-14}
     & turbulence2 &  &  &  &  &  &  &  &  &  &  &  & \\ 
    \cline{2-14}
     & turbulence3 &  &  &  &  &  &  &  &  &  &  &  & \\ 
     \hline
     
\end{tabular}}
\label{table:traintest1}
\end{table*}

\begin{table*}
\floatpagestyle{empty}
\centering
\caption{Training and test splits $\mathbf{S_{13}}$ to $\mathbf{S_{18}}$ used for evaluation.}
\scalebox{0.92}{
\begin{tabular}{|c|c|c|c|c|c|c|c|c|c|}
    \hline
     \textbf{category} & \textbf{video}  & $\mathbf{S_{13}}$& $\mathbf{S_{14}}$& $\mathbf{S_{15}}$& $\mathbf{S_{16}}$& $\mathbf{S_{17}}$& $\mathbf{S_{18}}$ \\

     \hline \hline 
    \multirow{4}{*}{baseline} & highway  & \color{blue}{Tr} & \color{blue}{Tr} & \color{blue}{Tr} & \color{blue}{Tr} & \color{blue}{Tr} & \color{blue}{Tr}\\ 
    \cline{2-8}
     & pedestrians  & \color{blue}{Tr} & \color{blue}{Tr} & \color{blue}{Tr} & \color{blue}{Tr} & \color{blue}{Tr} & \color{blue}{Tr}\\ 
    \cline{2-8}
     & office & \color{blue}{Tr} & \color{blue}{Tr} & \color{blue}{Tr} & \color{blue}{Tr} & \color{blue}{Tr} & \color{blue}{Tr}\\ 
    \cline{2-8}
     & PETS2006 & \color{blue}{Tr} & \color{blue}{Tr} & \color{blue}{Tr} & \color{blue}{Tr} & \color{blue}{Tr} & \color{blue}{Tr}\\

    \hline \hline 
    \multirow{4}{*}{\makecell{bad\\weather}} & skating  & \color{blue}{Tr} & \color{blue}{Tr} & \color{blue}{Tr} & \color{blue}{Tr} & \color{blue}{Tr} & \color{blue}{Tr}\\ 
    \cline{2-8}
     & blizzard & \color{blue}{Tr} & \color{blue}{Tr} & \color{blue}{Tr} & \color{blue}{Tr} & \color{blue}{Tr} & \color{blue}{Tr}\\ 
    \cline{2-8}
     & snowFall  & \color{blue}{Tr} & \color{blue}{Tr} & \color{blue}{Tr} & \color{blue}{Tr} & \color{blue}{Tr} & \color{blue}{Tr}\\ 
    \cline{2-8}
     & wetSnow  & \color{blue}{Tr} & \color{blue}{Tr} & \color{blue}{Tr} & \color{blue}{Tr} & \color{blue}{Tr} & \color{blue}{Tr}\\

    \hline \hline 
    \multirow{6}{*}{\makecell{intermittent\\object\\motion}} & abandonedBox  & \color{blue}{Tr} & \color{blue}{Tr} & \color{blue}{Tr} & \color{blue}{Tr} &  \color{blue}{Tr} & \color{blue}{Tr} \\ 
    \cline{2-8}
     & parking & \color{blue}{Tr} & \color{blue}{Tr} & \color{blue}{Tr} & \color{blue}{Tr} &  \color{blue}{Tr} & \color{blue}{Tr} \\ 
    \cline{2-8}
     & sofa  & \color{blue}{Tr} & \color{blue}{Tr} & \color{blue}{Tr} & \color{blue}{Tr} &  \color{blue}{Tr} & \color{blue}{Tr} \\ 
    \cline{2-8}
     & streetLight & \color{blue}{Tr} & \color{blue}{Tr} & \color{blue}{Tr} & \color{blue}{Tr} &  \color{blue}{Tr} & \color{blue}{Tr} \\ 
    \cline{2-8}
     & tramstop  & \color{blue}{Tr} & \color{blue}{Tr} & \color{blue}{Tr} & \color{blue}{Tr} &  \color{blue}{Tr} & \color{blue}{Tr} \\ 
    \cline{2-8}
     & winterDriveway & \color{blue}{Tr} & \color{blue}{Tr} & \color{blue}{Tr} & \color{blue}{Tr} &  \color{blue}{Tr} & \color{blue}{Tr} \\

    \hline \hline 
    \multirow{4}{*}{\makecell{low\\framerate}} & port 0.17fps  & \color{blue}{Tr} & \color{blue}{Tr} & \color{blue}{Tr} & \color{blue}{Tr} & \color{blue}{Tr} & \color{blue}{Tr}\\ 
    \cline{2-8}
     & tramCrossroad 1fps &  \color{blue}{Tr} & \color{blue}{Tr} & \color{blue}{Tr} & \color{blue}{Tr} & \color{blue}{Tr} & \color{blue}{Tr} \\ 
    \cline{2-8}
     & tunnelExit 0.35fps  & \color{blue}{Tr} & \color{blue}{Tr} & \color{blue}{Tr} & \color{blue}{Tr} & \color{blue}{Tr} & \color{blue}{Tr}\\ 
    \cline{2-8}
     & turnpike 0.5fps  & \color{blue}{Tr} & \color{blue}{Tr} & \color{blue}{Tr} & \color{blue}{Tr} & \color{blue}{Tr} & \color{blue}{Tr}\\

    \hline \hline 
    \multirow{6}{*}{shadow} & backdoor  & \color{blue}{Tr} & \color{blue}{Tr} & \color{blue}{Tr} & \color{blue}{Tr} & \color{blue}{Tr} & \color{blue}{Tr}\\ 
    \cline{2-8}
     & bungalows  & \color{blue}{Tr} & \color{blue}{Tr} & \color{blue}{Tr} & \color{blue}{Tr} & \color{blue}{Tr} & \color{blue}{Tr}\\ 
    \cline{2-8}
     & busStation & \color{blue}{Tr} & \color{blue}{Tr} & \color{blue}{Tr} & \color{blue}{Tr} & \color{blue}{Tr} & \color{blue}{Tr}\\ 
    \cline{2-8}
     & copyMachine & \color{blue}{Tr} & \color{blue}{Tr} & \color{blue}{Tr} & \color{blue}{Tr} &  \color{blue}{Tr} & \color{blue}{Tr} \\ 
    \cline{2-8}
     & cubicle & \color{blue}{Tr} & \color{blue}{Tr} & \color{blue}{Tr} & \color{blue}{Tr} &  \color{blue}{Tr} & \color{blue}{Tr} \\ 
    \cline{2-8}
     & peopleInShade & \color{blue}{Tr} & \color{blue}{Tr} & \color{blue}{Tr} & \color{blue}{Tr} & \color{blue}{Tr} & \color{blue}{Tr}\\

    \hline \hline 
    \multirow{4}{*}{\makecell{camera\\jitter}} & badminton &  \color{blue}{Tr} & \color{blue}{Tr} & \color{blue}{Tr} & \color{blue}{Tr} & \color{blue}{Tr} & \color{blue}{Tr}\\ 
    \cline{2-8}
     & traffic  & \color{blue}{Tr} & \color{blue}{Tr} & \color{blue}{Tr} & \color{blue}{Tr} & \color{blue}{Tr} & \color{blue}{Tr}\\ 
    \cline{2-8}
     & boulevard  & \color{blue}{Tr} & \color{blue}{Tr} & \color{blue}{Tr} & \color{blue}{Tr} & \color{blue}{Tr} & \color{blue}{Tr}\\ 
    \cline{2-8}
     & sidewalk  & \color{blue}{Tr} & \color{blue}{Tr} & \color{blue}{Tr} & \color{blue}{Tr} & \color{blue}{Tr} & \color{blue}{Tr}\\

    \hline \hline 
    \multirow{6}{*}{\makecell{dynamic\\background}} & boats &  \color{blue}{Tr} & \color{blue}{Tr} & \color{blue}{Tr} & \color{blue}{Tr} & \color{blue}{Tr} & \color{blue}{Tr}\\ 
    \cline{2-8}
     & canoe & \color{blue}{Tr} & \color{blue}{Tr} & \color{blue}{Tr} & \color{blue}{Tr} & \color{blue}{Tr} & \color{blue}{Tr}\\ 
    \cline{2-8}
     & fall  & \color{blue}{Tr} & \color{blue}{Tr} & \color{blue}{Tr} & \color{blue}{Tr} & \color{blue}{Tr} & \color{blue}{Tr}\\ 
    \cline{2-8}
     & fountain01  & \color{blue}{Tr} & \color{blue}{Tr} & \color{blue}{Tr} & \color{blue}{Tr} & \color{blue}{Tr} & \color{blue}{Tr}\\ 
    \cline{2-8}
     & fountain02  & \color{blue}{Tr} & \color{blue}{Tr} & \color{blue}{Tr} & \color{blue}{Tr} & \color{blue}{Tr} & \color{blue}{Tr}\\ 
    \cline{2-8}
     & overpass  & \color{blue}{Tr} & \color{blue}{Tr} & \color{blue}{Tr} & \color{blue}{Tr} & \color{blue}{Tr} & \color{blue}{Tr}\\

    \hline \hline 
    \multirow{6}{*}{\makecell{night\\videos}} & bridgeEntry   &  &  &  &  &  & \\ 
    \cline{2-8}
     & busyBoulvard  &  &  &  &  &  & \\ 
    \cline{2-8}
     & fluidHighway &  &  &  &  &  & \\ 
    \cline{2-8}
     & streetCornerAtNight &  &  &  &  &  & \\ 
    \cline{2-8}
     & tramStation &  &  &  &  &  & \\ 
    \cline{2-8}
     & winterStreet &  &  &  &  &  & \\

    \hline \hline 
    \multirow{4}{*}{PTZ} & continuousPan  & \color{blue}{Tr} & \color{red}{Test} &  &  &  & \\ 
    \cline{2-8}
     & intermittentPan & \color{blue}{Tr} & \color{red}{Test} &  &  &  & \\ 
    \cline{2-8}
     & twoPositionPTZCam & \color{red}{Test} & \color{blue}{Tr} &  &  &  & \\ 
    \cline{2-8}
     & zoomInZoomOut  & \color{red}{Test} & \color{blue}{Tr} &  &  &  & \\

    \hline \hline 
    \multirow{5}{*}{thermal} & corridor &  &  & \color{blue}{Tr} & \color{red}{Test} & \color{blue}{Tr} & \color{blue}{Tr}\\ 
    \cline{2-8}
     & diningRoom  &  &  & \color{blue}{Tr} & \color{red}{Test} & \color{blue}{Tr} & \color{blue}{Tr}\\ 
    \cline{2-8}
     & lakeSide &  &  & \color{red}{Test} & \color{blue}{Tr} & \color{blue}{Tr} & \color{blue}{Tr}\\ 
    \cline{2-8}
     & library &  &  & \color{red}{Test} & \color{blue}{Tr} & \color{blue}{Tr} & \color{blue}{Tr}\\ 
    \cline{2-8}
     & park &  &  & \color{red}{Test} & \color{blue}{Tr} & \color{blue}{Tr} & \color{blue}{Tr}\\

    \hline \hline 
    \multirow{4}{*}{turbulence} & turbulence0  &  &  & \color{blue}{Tr} &  \color{blue}{Tr} & \color{red}{Test} & \color{blue}{Tr}\\ 
    \cline{2-8}
     & turbulence1 &  &  & \color{blue}{Tr} & \color{blue}{Tr} & \color{red}{Test} & \color{blue}{Tr}\\ 
    \cline{2-8}
     & turbulence2 &  &  & \color{blue}{Tr} & \color{blue}{Tr} & \color{blue}{Tr} & \color{red}{Test}\\ 
    \cline{2-8}
     & turbulence3 &  &  & \color{blue}{Tr} & \color{blue}{Tr} & \color{blue}{Tr} & \color{red}{Test}\\ 
     \hline

\end{tabular}}
\label{table:traintest2}
\end{table*}

\end{document}